\documentclass[letterpaper]{article} 
\usepackage{aaai23}  
\usepackage{times}  
\usepackage{helvet}  
\usepackage{courier}  
\usepackage[hyphens]{url}  
\usepackage{graphicx} 
\usepackage{natbib}  
\usepackage{caption} 
\usepackage{algorithm}
\usepackage{algorithmic}

\usepackage{tabularx}

\usepackage{newfloat}
\usepackage{listings}
\DeclareCaptionStyle{ruled}{labelfont=normalfont,labelsep=colon,strut=off} 
\floatstyle{ruled}
\newfloat{listing}{tb}{lst}{}
\floatname{listing}{Listing}
\title{Topic Segmentation Model Focusing on Local Context}
\author {
    Jeonghwan Lee, Jiyeong Han, Sunghoon Baek, Min Song\footnote{Corresponding author.}\\
}
\affiliations {
    Yonsei University\\
    jeonghwan.ai@yonsei.ac.kr, jiyoung181@yonsei.ac.kr, sunghoon.baek@yonsei.ac.kr, min.song@yonsei.ac.kr
}

\usepackage{bibentry}

\begin{document}

\maketitle

\begin{abstract}
Topic segmentation is important in understanding scientific documents since it can not only provide better readability but also facilitate downstream tasks such as information retrieval and question answering by creating appropriate sections or paragraphs. In the topic segmentation task, topic coherence is critical in predicting segmentation boundaries. Most of the existing models have tried to exploit as many contexts as possible to extract useful topic-related information. However, additional context does not always bring promising results, because the local context between sentences becomes incoherent despite more sentences being supplemented. To alleviate this issue, we propose siamese sentence embedding layers which process two input sentences independently to get appropriate amount of information without being hampered by excessive information. Also, we adopt multi-task learning techniques including Same Topic Prediction (STP), Topic Classification (TC) and Next Sentence Prediction (NSP). When these three classification layers are combined in a multi-task manner, they can make up for each other's limitations, improving performance in all three tasks. We experiment different combinations of the three layers and report how each layer affects other layers in the same combination as well as the overall segmentation performance. The model we proposed achieves the state-of-the-art result in the WikiSection dataset.
\end{abstract}

\section{Introduction}

Nowadays, we can easily access vast amounts of scientific documents such as PubMed and Wikipedia. A lot of researchers are studying ways to effectively use these documents in areas like information retrieval (IR), question answering (QA) and search engine. However, applying previous IR models (or QA models) directly on these documents is impossible because most of them assume an input size of at most a paragraph while these documents consist of multiple paragraphs. 
Furthermore, extracting crucial parts of each document does not necessarily require the whole document to be used. For example, to search for similar papers on a topic of interest we can simplify the problem by calculating cosine similarity between sections of each document rather than full text to save resources.

These are where topic segmentation can be used. Topic segmentation divides a document into segments with respect to the topic coherence of each segment. A well-divided document according to the topics provides better readability, making it easier for the readers to find the desired information in the document. Most importantly, it can facilitate downstream-tasks such as IR and QA. 

Although most of the existing topic segmentation models take topic coherence into consideration when dividing a document, they don't undergo the process of classifying topic labels for each sentence, even when these topic labels are useful for inferring topic coherence. Most importantly, inspired by Neural Text Segmentation Model\unskip~\cite{1407796:24888797}, these models are designed to take block of text as input, which possibly hinders understanding local context of the input text\unskip~\cite{1407796:24889355}.

\begin{figure}
\centering
\includegraphics[width=\columnwidth]{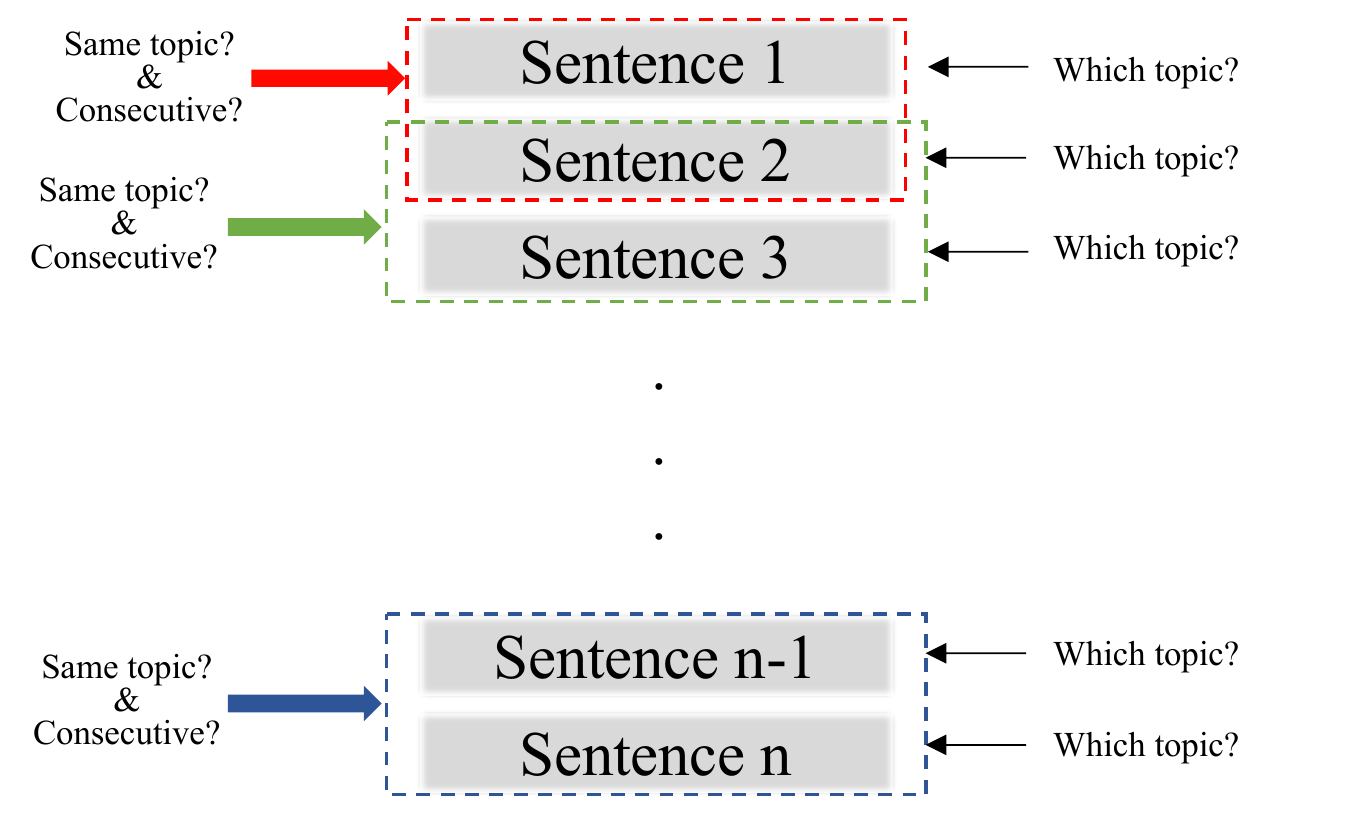}
\caption{A window with a size of 1 slides through the entire sentence, predicting the topic of each sentence. At the same time, the model determines whether the two sentences are in the same topic and whether the two sentences are consecutive.}
\label{figure_sgment}
\end{figure}

We try to tackle the above issues by adopting a siamese network to encode two input sentences independently and putting them through a multi-task learning algorithm that includes topic classification and other auxiliary tasks. First, in order to deal with two input sentences independently, we construct our model in a siamese network with sentence embeddings from a Sentence Transformer\unskip~\cite{1407796:24888985}. This method allows our model to preserve local context between the two input sentences without being overwhelmed by excessive information.

We consider topic segmentation as a Same Topic Prediction (STP) between two input sentences, following \unskip~\citet{1407796:24888889}. However, because our model processes only one sentence at a time to preserve its unique information, the model cannot observe context information across sentences. To alleviate this issue, we add two auxiliary tasks to capture local context information. One of them is Topic Classification(TC) which predicts the exact topic of the input sentences through a topic classification layer to assist STP with a detailed topic information. The other is a Next Sentence Prediction (NSP) layer\unskip~\cite{1407796:24888979}, which supports the model in understanding the relationship between consecutive sentences. Figure~\ref{figure_sgment} simply shows how our model works.

To sum up, our model deals with two input sentences independently via the siamese sentence embedding layer that preserves local context of input sentences. Also, we show that connecting tasks that utilize same input sentences to extract different features in the sentence in a multi-task manner improves topic segmentation performance. Consequently, our model achieves state-of-the-art in the topic segmentation task using the WikiSection dataset.

\section{Related Work}
\subsection{Topic Segmentation }\unskip~\citet{1407796:24888797} solved topic segmentation task as a supervised neural network model. Block of text that consists of several sentences is fed into the model and the model predicts whether each sentence should be a segmentation point.

\unskip~\citet{1407796:24914938} introduced $k$-sized left and right supporting sentences, where neighboring $k$ number of sentences support injecting context into input sentence. However,\unskip~\citet{1407796:24889355} pointed out that "local context" was more important than "global context" in topic segmentation task, implying that excessive context might decrease the performance. The information from various sentences can hinder predicting label of a single sentence due to deterioration in the model's understanding of local context.

\unskip~\citet{1407796:24914941} proposed Sector which includes a topic embedding layer in their architecture. This topic embedding layer is implemented for topic classification and the result of topic classification is then used for segmentation.

\unskip~\citet{1407796:24888889} treated topic segmentation as a Same Topic Prediction(STP) between two input paragraphs. STP determines whether two input paragraphs refer to the same topic. They also experimented diverse sampling methods, and among these methods we adopt consecutive sampling. Details about consecutive sampling will be explained at section~\ref{approach}.

\subsection{Sentence Embedding}Sentence embedding is a method of capturing the semantic relationships among words in a sentence \unskip~\cite{1407796:24918232}. Quality of sentence embedding is critical especially in the topic segmentation task, because the task inevitably has to capture as much information as possible from long sentences as well as short ones.\unskip~\citet{1407796:24888797} utilized Bi-LSTM to generate sentence embedding where word embedding vectors are extracted from Word2Vec with each word in a sentence as input, fed into the Bi-LSTM layer one by one, and the final sequence representation was made by max-pooling over the output of the LSTM.

After the introduction of BERT,\unskip~\citet{1407796:24888985} proposed Sentence BERT specialized in creating sentence embeddings. Sentence BERT is a fine-tuned version of BERT trained on NLI(Natural Language Inference) and STS(Semantic Textual Similarity) task. To handle input sentences effectively, the authors adopted siamese network which encodes each sentence independently and concatenates the encoded sentences to be fed into a classification layer. Consequently, Sentence BERT has better capability of dealing with long sentences, because the model understands high-level context of these sentences.

As another branch of Sentence BERT, SimCSE was proposed\unskip~\cite{gao-etal-2021-simcse}. The authors applied contrastive learning to forming sentence embeddings. They tried unsupervised method and showed that dropout could work as data augmentation and this prevented representation collapse. They also empirically and theoretically proved that contrastive learning objective was suitable for regularizing anisotropic space of a language model's embedding to be more uniform and it aligned positive pairs better in a supervised setting as a result. 

\unskip~\citet{1407796:24948221} proposed Cross-segment BERT. They used pre-trained BERT in which left and right context were separated via [SEP] token and encoded the sequence of word-piece tokens into sentence representations.\unskip~\citet{1407796:24888889} used Sentence BERT for a sentence encoder which is known to have substantial capability of understanding high-level context.

\section{Proposed Approach}\label{approach}
\subsection{Architecture}Our model follows the typical architecture of text segmentation models: a sentence embedding layer followed by a segment classifier, which is replaced by a Same Topic Prediction layer in our model.

However, our model takes two input sentences. To handle them independently, our model composes sentence embedding layer in siamese network form, so that the model receives an appropriate amount of information to predict the label. The encoded sentences are then fed into the topic classification layer one by one. By passing each sentence through the layer, the model acquires topic-related information of the sentence. Also, we adopt NSP layer to capture semantic relationship between the two sentences. Finally, STP layer predicts whether the sentences belong to the same topic.

We have $k$ documents $D_1,...D_k $ that $D $ consist of $n$ number sentences $s_1,...,s_n $, and the sentences are paired consecutively; $\lbrack(s_1,s_2),(s_2,s_3),...,(s_{n-2},s_{n-1}),(s_{n-1},s_n)\rbrack $. Each  $s_i(i\leq n) $ is assigned a topic label $t_i $ which describes topic label of $i $th sentence.

\begin{figure*}[!htbp]
\centering
\includegraphics[scale=0.8]{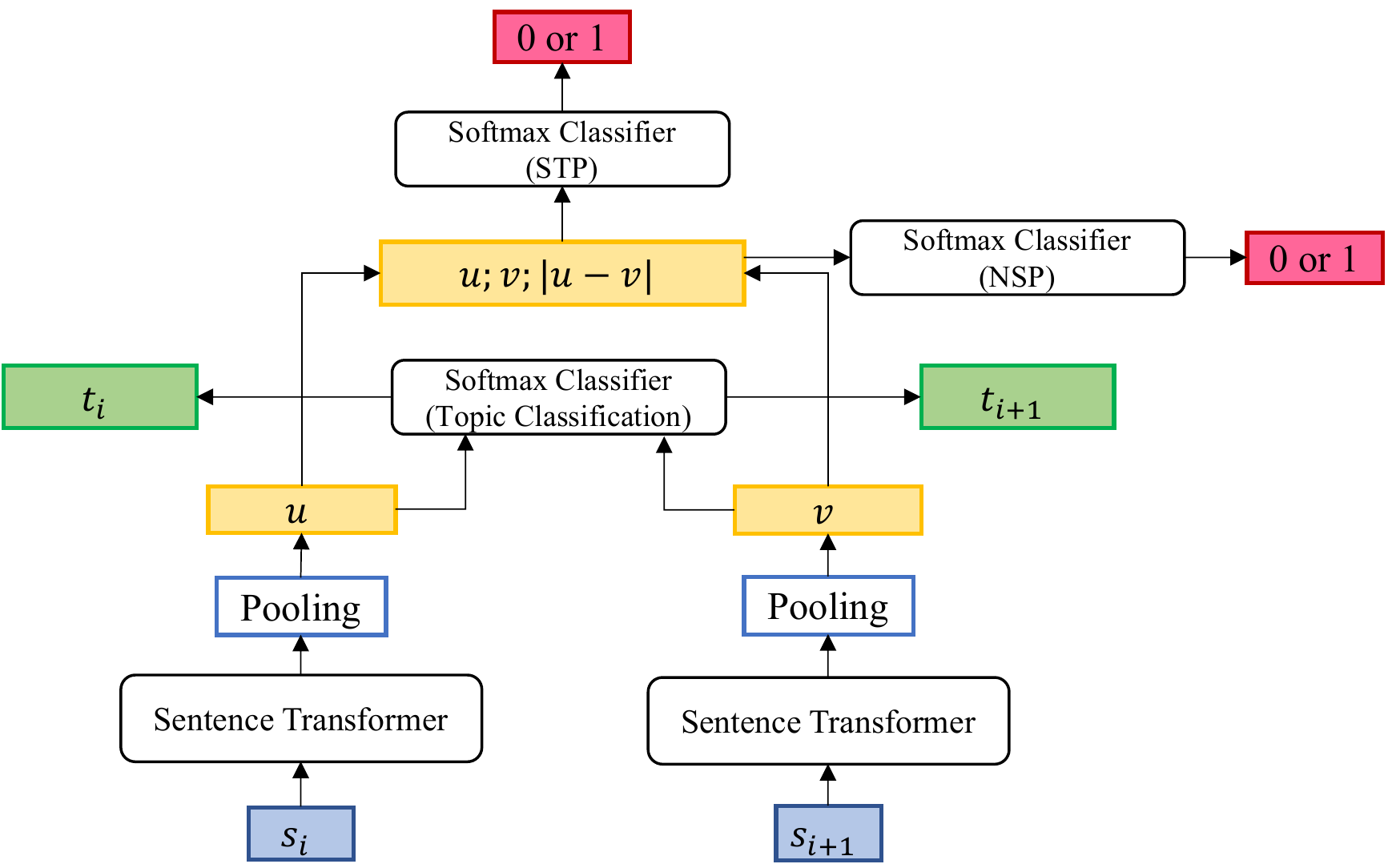}
\caption{The overview of our model. Two input sentences which are considered consecutive are fed into Sentence Transformer independently and encodes each input sentence. Each max-pooled encoded sentence, represented by $u $ and $v $ respectively, is fed into Topic classification layer. Before being fed into NSP layer and STP layer, we make concatenated feature $u;v;\vert u-v\vert $. Using $u;v;\vert u-v\vert $, NSP layer predicts if the two sentences are consecutive and STP layer finally determines whether they belong to same topic.}
\label{figure_architecture}
\end{figure*}

The sentence embedding layer encodes each input sentences $s_i\; $and $s_{i+1} $ and the encoded sentences are represented as $u $ and $v $, respectively. Figure~\ref{figure_architecture} shows the overview of our model.

\subsubsection{Siamese Sentence Embedding Layers from Sentence Transformer}
We propose siamese sentence embedding layer. In our model, Sentence Transformer encodes each sentence from two input sentences independently at the entry level. Then, the encoded sentences are concatenated before being fed into the STP layer. This method aims to preserve each sentence's unique information while acquiring local context between the two sentences.
\subsubsection{Multi Task Learning}
Our model has a total of three classification layers and we train them in a multi-task manner.

\textbf{Topic classification layer: }Topic classification layer is designed to capture exact topic information of a sentence. Topic classification is a multi-class classification that predicts the topic of an input sentence out of 30 labels for en\_city and 27 labels for en\_disease dataset\unskip~\cite{1407796:24914941} . This layer takes  $u $ and  $v $ one by one and predicts each topic label $t_i $ and $t_{i+1}$. 

\textbf{NSP layer: }NSP layer is fed with $u;v;\vert u-v\vert $\unskip~\cite{1407796:24888985} and the layer predicts NSP label.  This layer aims to supplement STP layer's limitation where STP layer can only determine whether the two input sentences are in the same topic and cannot determine if the sentences are actually consecutive. By adding NSP, the model can capture the semantic relationship between the two sentences, so the model can figure out whether the sentences are consecutive. NSP layer must go with consecutive sampling which will be explained below.

\begin{table}[tb!]
\centering
\begin{tabularx}{\columnwidth}{l>{\centering\arraybackslash}X>{\centering\arraybackslash}X>{\centering\arraybackslash}X}
\hline
\textbf{Models} & \textbf{STP Loss} & \textbf{TC Loss} & \textbf{NSP Loss} \\
\hline
\textbf{STP+TC} & 4 & 1 & - \\
\textbf{STP+NSP} & 1 & - & 1 \\
\textbf{STP+TC+NSP} & 4 & 1 & 4 \\
\hline
\end{tabularx}
\caption{Designated loss weights for each layer in case of multi-task learning}
\label{table_weights}
\end{table}

\textbf{STP layer: }STP layer is provided with    $u;v;\vert u-v\vert $ again and finally predicts segmentation label that is used to draw segmentation points in places where the two sentences belong to different topics.

\textbf{Multi task learning: }When the three layers are combined in a multi-task manner, they can make up each other's limitations. Since STP is based on binary classification, its task is much simpler than Topic Classification that is based on multi-class classification. However, since STP cannot capture the exact topic label of input sentences, Topic classification provides this information to the STP layer to help determine segmentation boundaries. The effect of NSP is explained above.

\textbf{Loss weight: }Because the losses from each layer are all different, there is a need to adjust the weights among the losses for improved model performance. We decide the weights for each loss by running numerous manual experiments and calculate the total loss using a weighted sum of the three losses from each classification layer. Table~\ref{table_weights} summarizes how each loss is weighted.

\subsection{Consecutive Sampling}
To make the model more robust, we add negative samples to the dataset by adopting consecutive sampling\unskip~\cite{1407796:24888889}. In consecutive sampling, all samples come from the same document. 

We have a document $D_a $ and a sentence $s_i^{t_i}\in D_a $ where the superscript $\begin{array}{l}t_i\\\end{array} $ refers to topic label of $s_i $. We pick one positive sample and two negative samples. The positive sample $s_{i+1}^{t_{i+1}=t_i}\in D_a $ is consecutive to $\begin{array}{l}s_i\\\end{array} $ and the first negative sample $s_{k\neq i+1}^{t_k=t_i} $ is from the same topic as $s_i $, but not consecutive to $\begin{array}{l}s_i\\\end{array} $. Finally, the second negative sample $s_l^{t_l\neq t_i} $ is from different topics, which is naturally considered not consecutive to $\begin{array}{l}s_i\\\end{array}$.

\section{Experiment}

\subsection{Dataset}
We use WikiSection\unskip~\cite{1407796:24914941} for training and evaluating our model. WikiSection covers two distinct domains: city and disease. Each domain has 19,539 and 3,590 documents, respectively, with various topics in each document. In total, there are 30 and 27 topics for each domain. The dataset is divided into 70\% training, 10\% validation and 20\% test sets. Table~\ref{table_numofdoc} gives statistics of the dataset.

\begin{table}
\centering
\begin{tabularx}{\columnwidth}{l>{\raggedleft\arraybackslash}X>{\raggedleft\arraybackslash}X}
\hline
\textbf{} & \textbf{en\_city} & \textbf{en\_disease} \\
\hline
\textbf{Docs} & 19,539 & 3,590 \\
\textbf{Topics} & 30 & 27 \\
\hline
\end{tabularx}
\caption{The number of documents and topics for \textbf{en\_city} and \textbf{en\_disease}}
\label{table_numofdoc}
\end{table}

\subsection{Experimental Setup}
We use nltk sentence tokenizer\footnote{NLTK :: Natural Language Toolkit} to split the documents into sentence units and apply consecutive sampling only on the training dataset. Table~\ref{table_numofrow} gives the data statistics after applying sentence split and consecutive sampling. We implement all-MiniLM-L12-v2 and from Sentence-Transformers\footnote{Pretrained Models {\textemdash} Sentence-Transformers documentation (sbert.net)} for our sentence encoder. We set the maximum epoch size to 14 but save the model only when the validation $P_k $ scores best. Batch size is 48, learning rate is $1e{-}6 $  and LinearLR scheduler is applied with the default parameters setting.

\begin{table}
\centering
\begin{tabularx}{\columnwidth}{l>{\raggedleft\arraybackslash}X>{\raggedleft\arraybackslash}X}
\hline
\textbf{} & \textbf{en\_city} & \textbf{en\_disease} \\
\hline
\textbf{Train} & 1,690,103 & 336,459 \\
\textbf{Valid} & 85,072 & 16,285 \\
\textbf{Test} & 168,924 & 31,110 \\
\hline
\end{tabularx}
\caption{The number of rows after applying nltk sentence tokenizer and consecutive sampling. Consecutive sampling is applied only on the trainset.}
\label{table_numofrow}
\end{table}

\begin{figure*}[htb!]
\centering
\includegraphics[width=\textwidth]{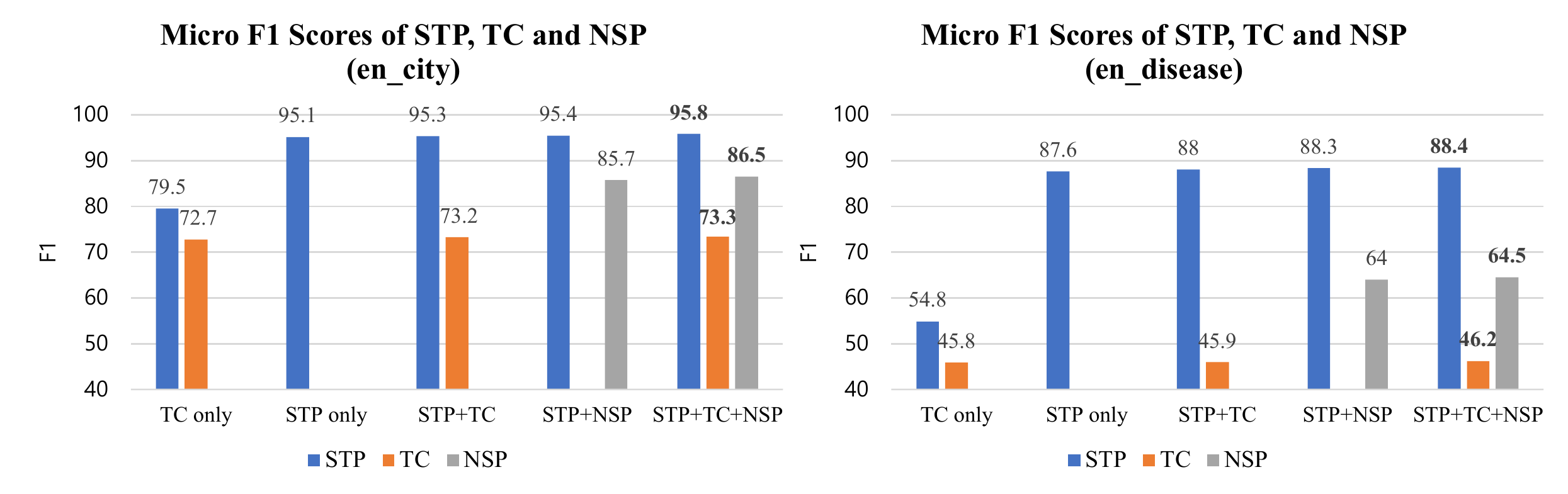}
\caption{Figure of F1 scores with combination of different task layers. In case of TC-only model, STP output is 1 if the results of topic classification on each sentence refer to the same topic otherwise 0.}
\label{figure_f1}
\end{figure*}

\begin{table*}[htb!]
\centering
\begin{tabularx}{\linewidth}{Xcccc}
\hline
\textbf{Dataset} & \multicolumn{2}{c}{\textbf{en\_city}} & \multicolumn{2}{c}{\textbf{en\_disease}} \\
\textbf{Metric} & $P_k$ & $WinDiff$ & $P_k$ & $WinDiff$ \\
\hline
SEC{\textgreater}T+emb & 15.5 & - & 26.3 & - \\
Transformer${}_{BERT}^{2} $ & 8.2 & - & 18.8 & - \\
BiLSTM + BERT & 9.3 & - & 28.0 & - \\
Cross-segment BERT n\_context = 2 & 15.4 & 27.4 & 33.9 & 59.0 \\
Cross-segment BERT n\_context = 4 & 18.3 & 32.2 & 34.8 & 60.3 \\
Cross-segment BERT n\_context = 6 & 45.1 & 50.0 & 34.0 & 57.1 \\
\hline
TC-only & 15.0 & 17.8 & 41.5 & 45.4 \\
STP-only & 5.1 & 5.8 & 14.8 & 15.8 \\
STP + TC & 5.0 & 5.7 & 14.0 & 15.0 \\
STP + NSP & 4.9 & 5.6 & 14.1 & 15.1 \\
\textbf{STP + TC + NSP} & \textbf{4.6} & \textbf{5.2} & \textbf{13.7} & \textbf{14.7} \\
\hline
\end{tabularx}
\caption{Test $P_k$ and $WindowDiff$ scores of baseline models and our models. Note that the $WinDiff$ metric is used only in our models and the Cross-segment BERT models. We reimplement Cross-segment BERT ourselves following their official codes.}
\label{table_score}
\end{table*}

\subsection{Metric}
For a comprehensive evaluation, we used \textbf{$P_k $}, $WindowDiff $ and micro F1 score to evaluate our models. We use $P_k $ score for making comparisons between our models and all other baseline models, and $WindowDiff $ is used to evaluate ours and Cross-segment BERT that we implemented. F1 score is used for the purpose of ablation study on our models.

\subsubsection{$P_k $}
$P_k $\unskip~\cite{1407796:25052335} is a probability that a segmentation model performs an incorrect segmentation. While a sliding window of size k passing over the sentences, the status (0 or 1) is determined by whether the two ends of the window are in the same segment or in different segments. $P_k $ is calculated by counting unmatched cases between the ground truths and predicted values. As in many previous studies, we set the window size k to half the average segment length of the ground truths.

\subsubsection{$WindowDiff $}
$WindowDiff $\unskip~\cite{1407796:25053072} is an improved metric from $P_k $ in that it alleviates the impact of false negative penalty and segment size distribution.

Similar to $P_k $, $WindowDiff $ score also uses sliding window and compares the ground truths with the predicted values. However, this metric also takes the number of boundaries into consideration. It is closer to the ground truth when the models get a lower score in both $P_k $ and $WindowDiff $.

\subsection{Baseline Models}
We compare our model with competitive neural text segmentation baselines 1) SEC{\textgreater}T+emb\unskip~\cite{1407796:24914941}, 2) Transformer${}_{BERT}^{2} $\unskip~\cite{1407796:24933294}, a framework based on two transformers, where one is a pre-trained transformer for encoding sentences and the other is a transformer for segmentation, 3) Bi-LSTM + BERT\unskip~\cite{1407796:24889355}, that is based on a hierarchical attention Bi-LSTM network, and 4) Cross-segment BERT\unskip~\cite{1407796:24948221}, which handles left and right context simultaneously using a BERT encoder.

We adopt the results of SEC{\textgreater}T+emb from \unskip~\citet{1407796:24914941}, Transformer${}_{BERT}^{2}$ and Bi-LSTM + BERT from \unskip~\citet{1407796:24889355}. We implement Cross-segment BERT ourselves following their official code while applying diverse size of context.

\begin{figure*}[htb!]
\centering
\includegraphics[scale=0.8]{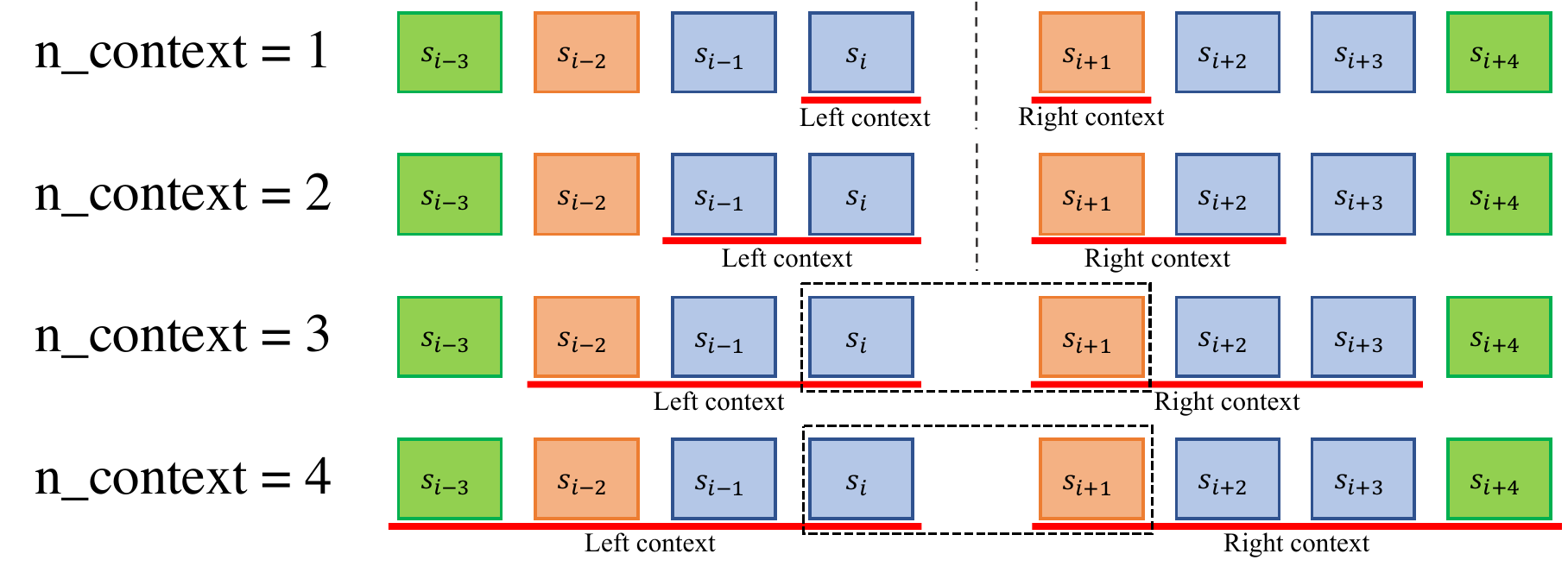}
\caption{Effect of context size on prediction. The color of the box represents the topic of the sentence and the red line represents supporting context. The vertical dotted line represents a segmentation point between $s_i $ and $s_{i+1} $ while the dotted box describes that the two sentences are not segmented. $\begin{array}{l}s_i\\\end{array} $ and $\begin{array}{l}s_{i+1}\\\end{array} $ should be divided, since they belong to different topics, but are not segmented in cases of $n\_context\;=3 $ and $\begin{array}{l}n\_context\;=4\\\end{array} $.}
\label{figure_ncontext}
\end{figure*}

\subsection{Results and Analysis}

We report evaluation results on Figure~\ref{figure_f1} and Table~\ref{table_score}. Figure~\ref{figure_f1} summarizes how combination of each classification layer affects their F1 scores. Table~\ref{table_score} shows performance comparison between our model and other baseline models in $P_k $ and $WindowDiff $, respectively. Our proposed models, except for TC-only model, outperform all the baseline models by a large margin.

\subsubsection{\textbf{Effect of MTL}}
Figure~\ref{figure_f1} shows F1 scores derived from combinations of tasks mentioned above. We can see that MTL is effective in improving the performance, which applies to not only the performance of STP that is responsible segmentation but also the performances of TC and NSP. This is believed to be because, as we pointed out in the section~\ref{approach}, the layers make up for each other's limitations by extracting different features from same input sentences that assist understanding semantic information.

\subsubsection{\textbf{STP-only vs TC-only}}
In order to verify the effectiveness of STP layer, we also experiment TC-only model, which is close to Sector in that segmentation is performed only using topic labels.

$P_k$ and $WindowDiff$ of TC-only model are much higher than those of STP-only model. Poor classification performance of Topic classification directly causes this phenomenon. Figure~\ref{figure_f1} indicates that F1 scores of TC-only model are significantly lower than those of STP-only. Because topic classification layer is based on multi-class classification, which is more difficult than the binary classification of STP-only. 

\subsubsection{\textbf{STP vs NSP}}
Although STP and NSP have the same architecture, STP's F1 scores are always higher than NSP's in both datasets. We assume that this difference is derived from the difference in the information that STP and NSP focus on. STP-only determines whether the two sentences belong to the same topic, so it only pays attention to topic differences between two sentences. In other words, due to the nature of the task, STP does not consider the relationship between two input sentences. However, in the NSP task, the layer faces difficulties as two sentences may not be consecutive even if they belong to the same topic because of our consecutive sampling. Thus, NSP must find the semantic relationship between the two sentences as well as topic coherence, which makes the task tricky.

\subsubsection{\textbf{How the number of contexts affects the performance}}
To show the importance of local context, we implement Cross-segment BERT\unskip~\cite{1407796:24948221} by applying diverse size of context on the model. Table~\ref{table_score} shows that raising context size rather deteriorates the performance. We conjecture that this is because the more sentences there are, the more likely for different topics to be mingled, which likely interferes the model from understanding local context with overflow of noise. Because processing multiple sentences simultaneously using a left and right context structure rather adds noise to the contexts, we choose to encode the two input sentences independently.

Figure~\ref{figure_ncontext} explains this local context capturing error. The models in Figure~\ref{figure_ncontext} are all expected to create a segmentation point between $\begin{array}{l}s_i\\\end{array}$ and $\begin{array}{l}s_{i+1}\\\end{array}$, but models with larger context sizes fail to split the two sentences, because segmentation only takes into account the overall context of each side. As the context size increases, the model suffers from generalization and interprets left and right contexts as similar even when the two specific sentences refer to different topics and hence should be segmented.

Also, Cross-segment BERT encodes left and right context simultaneously. Because Cross-encoder inevitably makes context of one input sentence influence the other\unskip~\cite{1407796:25045363}, the unique information of each context can change unexpectedly. Therefore, we process each sentence independently via siamese sentence embedding in order to preserve the original local context.

\subsubsection{\textbf{Dealing with Scientific Documents}}
We can also find that the scores for en\_disease are underperforming compared to that for en\_city. We assume that this result due to the fact that en\_disease is more science domain specific (i.e. biology) while en\_city covers relatively general topics. \unskip~\citet{1407796:24914941} commented that documents in en\_disease are described in a precise language, but on the other hand those in en\_city are described in a common language.
Considering that our backbone, miniLM was pre-trained on general documents like Wikipedia, the result seems natural.

To improve the performance on en\_disease, we implement SPECTER\unskip~\cite{cohan-etal-2020-specter} which was trained on scientific papers using Sci-BERT as the backbone model.

As shown in table~\ref{table_specter}, $P_k$ and $WinDiff$ improved by 1.6 and 1.8 respectively in en\_disease compared to the miniLM based model. We attribute the improvement to SPECTER's understanding of scientific documents. We expect the scores for en\_disease to be improved if we use more biology domain specific model like BioBERT as the backbone.

Interestingly, although the number of parameters of SPECTER was twice as large as that of miniLM (i.e. 768 vs 384), there was no improvement in the performance in en\_city. From this result, we can again confirm that domain knowledge is critical to the performance.

\begin{table}
\centering
\begin{tabularx}{\columnwidth}{l>{\raggedleft\arraybackslash}X>{\raggedleft\arraybackslash}X}
\hline
\textbf{} & \textbf{$P_k$} & \textbf{$WinDiff$} \\
\hline
\textbf{en\_city} & 4.6 & 5.2 \\
\textbf{en\_disease} & 12.1 & 12.9 \\
\hline
\end{tabularx}
\caption{Test $P_k$ and $WindowDiff$ scores of SPECTER based Topic Segmentation Model with STP+TC+NSP}
\label{table_specter}
\end{table}

\section{Conclusion and Future Work}
In this work, we propose our topic segmentation model which consists of siamese sentence embedding layer from Sentence Transformer and three classification layers. With several different experiments, we show that our proposed model outperforms all the existing models. We also find that combining Same Topic Prediction, Topic Classification and Next Sentence Prediction in a multi-task manner increases segmentation performance.

Moreover, we empirically show the importance of local context in topic segmentation task. Contrary to the popular belief, increasing the number of context can rather degrade the performance due to generalization of local context. Our experiment indicates that narrowing context through our siamese sentence embedding layer can be effective in preserving local context.

Future work can highlight on the theoretical approach to local context. Although we empirically showed the influence of context size to the model performance in this paper, we did not concentrate on how we can determine which input sentences can provide substantial information in performing segmentation tasks. If we can infer each sentence's significance in prediction, we expect the model to capture the important sentences autonomously, consequently making the model agnostic to the context size.

\appendix

\section{Acknowledgments}
This work was supported by Institute of Information \& communications Technology Planning \& Evaluation (IITP) grant funded by the Korea government(MSIT) (No. 2020-0-01361, Artificial Intelligence Graduate School Program (Yonsei University)).

\bibliography{aaai23.bib}


\end{document}